\documentclass[letterpaper, 10 pt, conference]{ieeeconf}  

\IEEEoverridecommandlockouts                              

\usepackage{graphicx} 
\usepackage{amsmath} 
\usepackage[table]{xcolor}
\usepackage{tabularx}
\usepackage{booktabs}
\usepackage{dcolumn}
\usepackage{cite}
\graphicspath{{./media/}}

\usepackage{url}

\title{\LARGE \bf
A Multi-Layer Sim-to-Real Framework for \\Gaze-Driven Assistive Neck Exoskeletons}

\author{Colin Rubow$^{2,3,\ddagger}$,
Eric Brewer$^{1,\ddagger}$, Ian Bales$^{2}$, Haohan Zhang$^{2,3}$, and Daniel S. Brown$^{1,3,*}$
\thanks{This work was partially supported by the National Institutes of Health (R21EB035378). Colin Rubow also received support from the National Institute of Occupational Safety and Health (5T42OH008414).}
\thanks{$^{\ddagger}$These authors contributed equally.}
\thanks{$^{1}$Kahlert School of Computing, University of Utah, UT 84112, USA.}%
\thanks{$^{2}$Dept. of Mechanical Engineering, University of Utah, UT 84112, USA.}%
\thanks{$^{3}$Robotics Center, University of Utah, UT 84112, USA.}%
\thanks{*Corresponding Author (daniel.s.brown@utah.edu).}%
}

\begin{document}

\maketitle
\thispagestyle{empty}
\pagestyle{empty}

\begin{abstract}
Dropped head syndrome, caused by neck muscle weakness from neurological diseases, severely impairs an individual’s ability to support and move their head, causing pain and making everyday tasks challenging. Our long-term goal is to develop an assistive powered neck exoskeleton that restores natural movement. However, predicting a user’s intended head movement remains a key challenge. We leverage virtual reality (VR) to collect coupled eye and head movement data from healthy individuals to train models capable of predicting head movement based solely on eye gaze. We also propose a novel multi-layer controller selection framework, where head control strategies are evaluated across decreasing levels of abstraction—from simulation and VR to a physical neck exoskeleton. This pipeline effectively rejects poor-performing controllers early, identifying two novel gaze-driven models that achieve strong performance when deployed on the physical exoskeleton. Our results reveal that no single controller is universally preferred, highlighting the necessity for personalization in gaze-driven assistive control. Our work demonstrates the utility of VR-based evaluation for accelerating the development of intuitive, safe, and personalized assistive robots.
\end{abstract}

\section{INTRODUCTION}
Dropped Head Syndrome (DHS) is characterized by the inability of a person to move and raise their head due to neck weakness~\cite{lange1986floppy, drain2019dropped}. DHS can result from multiple causes, such as central/peripheral neurological pathology (e.g., amyotrophic lateral sclerosis) and autoimmune conditions (e.g., polymyositis)~\cite{suarez1992dropped, katz1996isolated, finsterer2010steroid, brodell2020dropped}. DHS causes fatigue during head movement and, in severe cases, results in a ``head-on-chest'' posture~\cite{brodell2020dropped}. This posture makes it challenging to breathe, swallow, and speak due to the bent airway. It also interferes with social interactions and walking~\cite{kiernan2011amyotrophic, walling1999amyotrophic}.

Static neck braces are currently used as a treatment for DHS. However, patients find these braces uncomfortable and ineffective~\cite{burke2025cervical, pancani2016assessment}. These braces fix the head in a static pose using support under the chin, making it difficult to open the mouth to speak or swallow. More importantly, these braces do not promote mobility necessary for daily tasks (e.g., feeding, horizontal vision, etc.)~\cite{miller2010soft}. Other solutions include using a reclining wheelchair with a headrest or fixing the head to a chair using straps. These solutions, however, are not portable and similarly do not allow head-neck motions.

Wearable robots (e.g., powered neck exoskeletons) that can restore head-neck movement through external actuators represent a potentially transformative solution to treat DHS. 
Although there have been significant advances in developing the mechanical hardware of these devices~\cite{zhang2017active, torrendell2024neck, kulkarni2024design, bales2024kinematic, demaree2023structurally, lingampally2019kinematic}, controlling these devices by a user remains a challenging problem. Using a hand-held device (e.g., joystick, keyboard) to control the neck exoskeleton is unintuitive and even infeasible for many patients, especially those whose DHS is a result of widespread neural degeneration (e.g., amyotrophic lateral sclerosis) and thus suffer from hand weakness~\cite{zhang2022amyotrophic, demaree2024preliminary}.

In contrast to hand-held devices, we propose and evaluate models that use a user's gaze as the control input for wearable neck robots. This is biologically motivated and viable because the eyes and the head often coordinate during visual tasks through neural mechanisms such as the vestibulo-ocular reflex (VOR). In VOR, compensatory rotations of the eyes allow the eyes to remain fixated on a target while the head moves towards the target~\cite{tweed1995eye, land2006eye}. Beyond VOR, it has also been found that the peak head velocity during gaze shifts is linearly proportional to the distance the head will travel~\cite{freedman2008coordination, freedman1997eye} and generally the head will travel directly towards the gaze target~\cite{tweed1995eye, tweed1997three}. Furthermore, it has been shown that head motions are minimized when the gaze lies in a central region in the field of view~\cite{freedman2008coordination} for local visual exploration, which we term the deadzone. Due to these behaviors, using eye movements as control inputs for wearable neck robots to restore head movement is promising.

\begin{figure}[t]
    \centering
    \includegraphics*[width=0.73\linewidth]{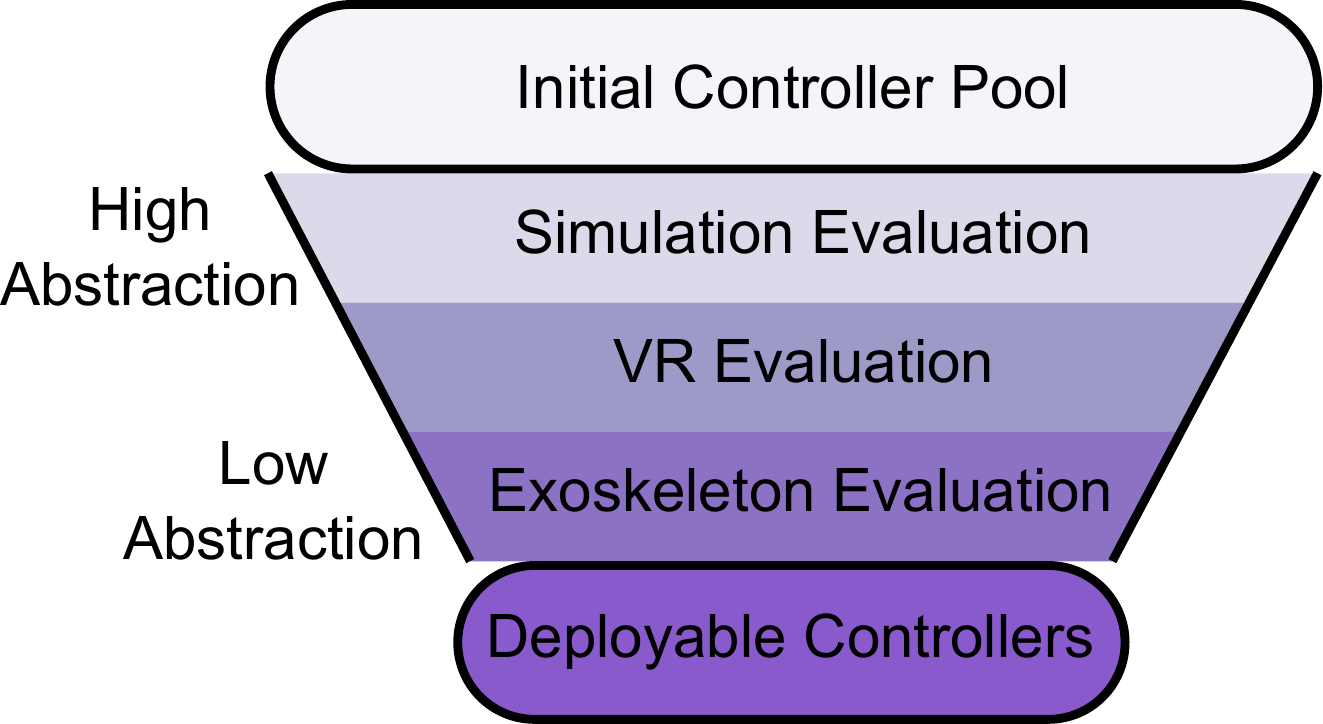}
    \caption{We propose to leverage multiple levels of abstraction and emulation to assist in efficiently finding useful gaze-driven neck exoskeleton controllers. As the initial controller pool moves down through each layer, poor controllers are rejected from further study. Our goal is to arrive at a final set of controller designs suitable for restoring head-neck motions. \label{fig:funnel}}
\end{figure}

During early development of these controllers, the question arises as to how to evaluate them. Deploying each controller on a physical robot with a human subject is time-consuming and could result in safety hazards. We therefore employ a novel, multi-level controller selection technique that uses three layers of abstraction to eliminate poor-performing controllers early in the development process (Fig.~\ref{fig:funnel}). The first layer tests the ability of the models to predict head orientations during eye-head coordination tasks taken from the previous dataset. The second deploys the controllers in a safe VR environment to assist healthy human subjects in performing eye-head coordination tasks without actually applying forces to the head. The third deploys the controllers on a physical neck exoskeleton to assist human subjects to perform a real-world eye-head coordination task.

We employ two strategies for identifying gaze-driven control models. The first is to learn the parameters of an analytical, hand-crafted model based on eye-head coordination behavior described in the literature~\cite{freedman1997eye}. The second strategy is fully data-driven, where we train multiple neural network models of varying architectures. These models take eye position and head orientation as inputs and predict the next desired head orientation. We train both styles of model using a large dataset of coupled eye and head movement data we collected using virtual reality (VR)~\cite{rubow2024dataset}.  

One of our main contributions is a multi-layer controller evaluation framework (see Fig.~\ref{fig:funnel}) that enables safe, efficient rejection of poor exoskeleton controllers before physical deployment.
With the goal of developing more natural eye-tracking controllers, we developed seven candidate controllers and evaluated them with our selection strategy. We were able to reject five of these controllers before physical robot deployment; and the other two controllers were deployed and approved by subjects in our user study.


\section{CONTROLLER MODELS \label{sec:controllers}}
Four categories of controllers were used in this work. 

\subsection{Quadrant (Baseline)}
This controller divides the range of view into four directional regions (up, down, left, right) and a central deadzone region of $10^\circ$ diameter~\cite{chang2020bio}. If the gaze lies in one of the directional regions, the controller moves the head at a constant speed in that direction at $20^\circ/$sec. If in the deadzone, the controller does not move. The controller only performs motions in purely horizontal or vertical directions and thus performs diagonal trajectories in a zig-zag fashion. While this controller employs VOR~\cite{tweed1995eye}, the speed and deadzone dimensions were not biologically chosen, and the zig-zag motions can be unnatural. However, to the best of our knowledge, this is the only eye controller developed for neck exoskeletons currently, so we use this controller as a baseline. In the following sections, we discuss our novel learned gaze-driven controllers.

\subsection{Vector Parameterized}
From literature~\cite{tweed1995eye, land2006eye, freedman2008coordination, freedman1997eye, tweed1997three}, we identified three key elements of eye-head coordination: VOR, peak head velocity being proportional to head displacement, and gaze deadzone. From these elements, we developed the following control law:

\begin{equation*}
    h(x) = \frac{v(x - c)}{1 + e^{a(-(x - c) + b)}}.
\end{equation*}

\begin{figure}[t]
    \centering
    \includegraphics*[width=0.9\linewidth]{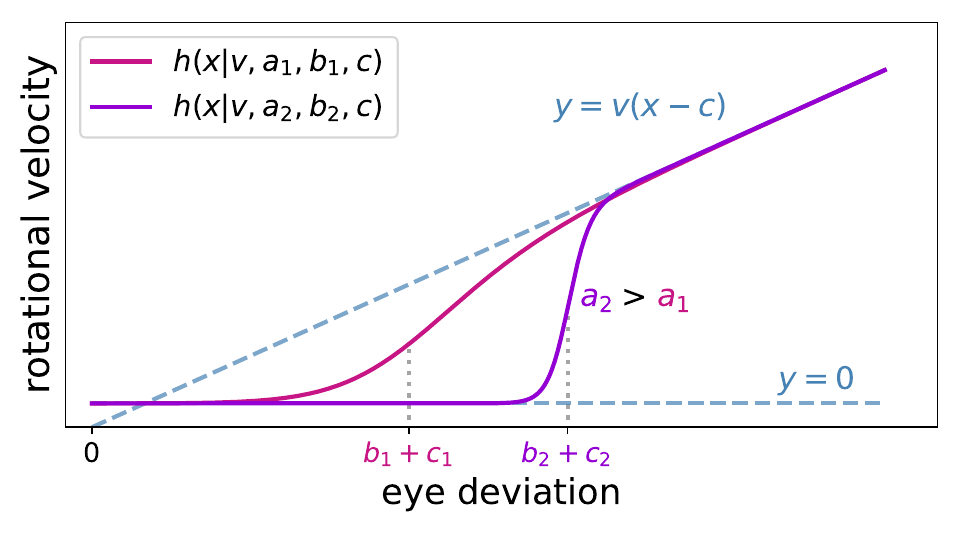}
    \caption{Two examples of the Vector Parameterized control law for either pitch or yaw directions. The $v$ and $c$ values are equal for both examples. The control law transitions from a deadzone with zero velocity to a line with proportional velocity. Gray dotted lines mark inflection point locations of the plots. Blue dashed lines show what the control law transitions to and from. The Purple curve on the right has a larger but stricter deadzone. \label{fig:control_law}}
\end{figure}

This function takes a gaze angle $x$ in a single dimension, i.e., pitch or yaw, and produces an angular rotation of the head in the same dimension. As $x$ increases, $h$ asymptotically approaches the linear function $f(x) = v(x - c)$ (two examples are shown in Fig.~\ref{fig:control_law}). This function induces a ``soft dead zone'' such that $h(x) \approx 0$ when $x \approx 0$ (i.e., the head remains still when the user is looking forward).
The parameter $v$ represents the linear asymptote slope, $1/a$ the exponential time constant and thus the ``sharpness'' of the inflection between the ``dead zone'' and asymptote, $b$ the location in terms of $x$ of the inflection point, and $c$ the shift of the function from the origin.

\subsection{Multi Layer Perceptron Model (MLP)}
Our Vector Parameterized model is based on key insights from eye-head coordination; however, it follows a specific functional form and its parameters are fit to multiple subjects' data, so the model could be sub-optimal. We thus train some simple feedforward MLP models on eye-head coordination data. We posited that a data-driven neural network would be able to induce eye tracking controllers that perform better than the analytical model we developed. As input, the model takes the 3D normalized direction of the left eye relative to the orientation of the head, that of the right eye, and that of the head itself relative to the environmental space. The MLP outputs the relative change of the head direction in the next time step in terms of angular pitch and yaw.

\subsection{Long Short Term Memory Model (LSTM)}
Due to the sequential nature of gaze and head vector data as a series over time, we also chose to evaluate the efficacy of using an LSTM~\cite{hochreiter1997long}, a popular recurrent neural network architecture, as a gaze-driven controller. An LSTM network has an inherent advantage in that it can theoretically capture inferential context over a large number of time steps. This means that the LSTM has the capability to model, for example, if a certain sequence of directional gaze vectors some number of time units in the past is currently relevant to predicting the motion of the head. In essence, the LSTM is capable of modeling eye-head coordination in the case that this phenomenon is not totally a Markov process, i.e., that its prediction depends only on the most recent time step. It takes as input the current gaze and head directions as 3D vectors. The LSTM retains a latent hidden state vector, which enables the capturing of context over time as described above.

\subsection{Training Details} The Parameterized Vector, MLP, and LSTM models were trained using the Adam optimizer~\cite{kingma2014adam}. The parameters were optimized by minimizing the mean-squared error between those models' predicted angular head movements and the actual angular head movements in the Eye Tracking VR Dataset~\cite{rubow2024dataset}. This dataset contains eye and head directional data for 25 participants (over 120 thousand data samples); each participant performed twelve 90-second trials (three 90-second trials for each of four distinct tasks) at 90 frames per second. We used data from 18 of the 25 participants for model training. After training these models, we evaluated them against a held-out test set consisting of data from the remaining 7 participants.

In practice, we had to downsample the input frequency for each LSTM variant by a factor of 2. This was because the ensuing VR experiment was only able to perform inference on LSTM-variant models at a rate of about 45~Hz, although the game itself runs at about 90~Hz. Thus, our LSTM models were trained using trajectories of length 4050 (90~s $\times$ 45~Hz). Other models did not experience any unintentional slowdown in terms of latency and were trained using trajectories containing 8100 time steps.

\section{AUTOREGRESSIVE EXPERIMENT}
As a result of training, our models were optimized sufficiently in terms of MSE loss, as shown in Table~\ref{tab:autoreg_table}. Nonetheless, we conducted an additional experiment in an attempt to simulate whether these controllers would behave well autoregressively when deployed with real human subjects.

\subsection{Methods -- Autoregressive}
We assume that a participant's gaze would approximately follow the same path throughout a given trial, \textit{regardless of} the underlying model controlling the rotation of the head. For example, in a Pursuit task, we assume that a participant would roughly look at the same sequence of points in 3D space (corresponding to the position of the object being tracked), irrespective of the actual rotation of their head.

Prior to the experiment, we calculated the sequence of \textbf{focal points} of each participant's binocular gaze in a given trial. During each time step within our dataset, we recorded the \textit{position} and \textit{orientation} in 3D space of the head and the \textit{position} and \textit{direction} of the eyes. Given these additional data, we then defined the gaze rays (each specified by an origin and direction) for both eyes. We can then calculate the focal point as the point in space closest to the gaze rays of both eyes. This is done by first taking the cross product of the directional vectors of both eyes,
    $\vec{g}_f = \vec{g}_l \times \vec{g}_r$,
\noindent where $\vec{g}_l$ and $\vec{g}_r$ are the normalized direction vectors of the left and right eyes. This yields $\vec{g}_f$, the direction of the only orthogonal, hence shortest, segment between the rays. It is then possible to solve via Gaussian elimination the following:
\begin{align*}
    \vec{p}_l + c_l * \vec{g}_l + c_f * \vec{g}_f &= \vec{p}_r + c_r * \vec{g}_r \\
    c_l * \vec{g}_l - c_r * \vec{g}_r + c_f * \vec{g}_f &= \vec{p}_r - \vec{p}_l
\end{align*}
\noindent where $\vec{p}$ is the origin of a particular ray, $c$ is the magnitude of that ray, and $\vec{g}$ is the direction. Solving the above system produces the magnitudes $c$ for the left-eye ray, right-eye ray, and the focal segment. We defined the focal point as the midpoint of that focal segment.

For most instances in our dataset, the rays representing the gaze of the left and right eyes nearly intersected a few meters in front of the participant, as expected. However, in the case when the user was blinking, we linearly interpolate the gaze direction between the two successful recordings surrounding the blinking interval, then normalize.


Given a participant's sequence of estimated focal points throughout a trial, the experiment was conducted as follows. First, given a single model from Section~\ref{sec:controllers}, we maintained the location and orientation of a virtual ``head'' within a 3D coordinate space. We also kept track of the location of the pupils relative to the head. We then applied our assumption---that the eyes would approximately look at the calculated focal point for this time step. Instead of setting the direction of both eyes to the ground truth directions, we aimed each eye toward the focal point, subject to a small amount of Gaussian noise. Next, we fed the selected model with the novel assumed gaze. This produces both a pitch and yaw value, which we used to rotate the virtual environment. We then repeat this process through the entirety of a given 90-second trial. Essentially, we perform a simulated rollout of a given trial in which the eyes are approximately ``locked on'' to the ground-truth \textbf{focal point} while the head rotates at the behest of the controller.

\subsection{Results -- Autoregressive}
\begin{figure}[htb]
    \centering
    \includegraphics[width=0.95\linewidth]{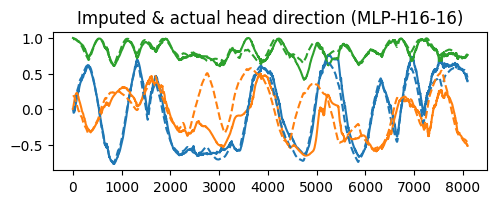}
    \includegraphics[width=0.95\linewidth]{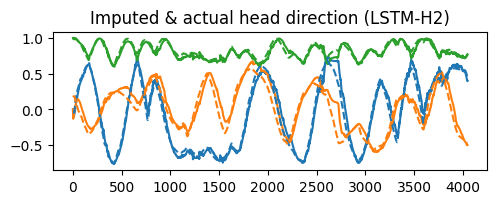}
    \includegraphics[width=0.95\linewidth]{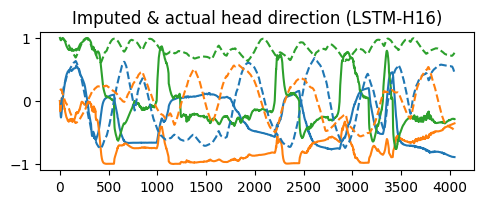}
    \caption{Trajectories of an autoregressive rollout for one MLP and two LSTM variations. The y-axis shows the $x$ (blue), $y$ (orange), and $z$ (green) values of the normalized direction of the head where $(0, 0, 1)$ represents the forward direction. The x-axis shows the time step $t$ throughout the trajectory. Solid lines represent the direction of the simulated, rotated ``virtual head'' during the experiment. Dashed lines represent the ground-truth direction of the head as seen in the dataset. (\textbf{Top}) An MLP with two intermediate hidden layers each of size \texttt{16}. (\textbf{Middle}) An LSTM with a context vector of size \texttt{2}. (\textbf{Bottom}) An LSTM with a context vector of size \texttt{16}.}
    \label{fig:autoreg_traj}
\end{figure}

We show in Fig.~\ref{fig:autoreg_traj} sample trajectories of three different models produced by the autoregressive routine. In the first two plots---those of MLP-H16\_16 and LSTM-H2---the imputed head direction roughly follows the actual head direction. The third plot, however, shows that the head direction deviated dramatically from its true course. \textit{Qualitatively}, this indicates that this controller will exhibit undesirable behavior in the VR experiment or that of the robot.

\begin{figure}[htb]
    \centering
    \includegraphics[width=1\linewidth]{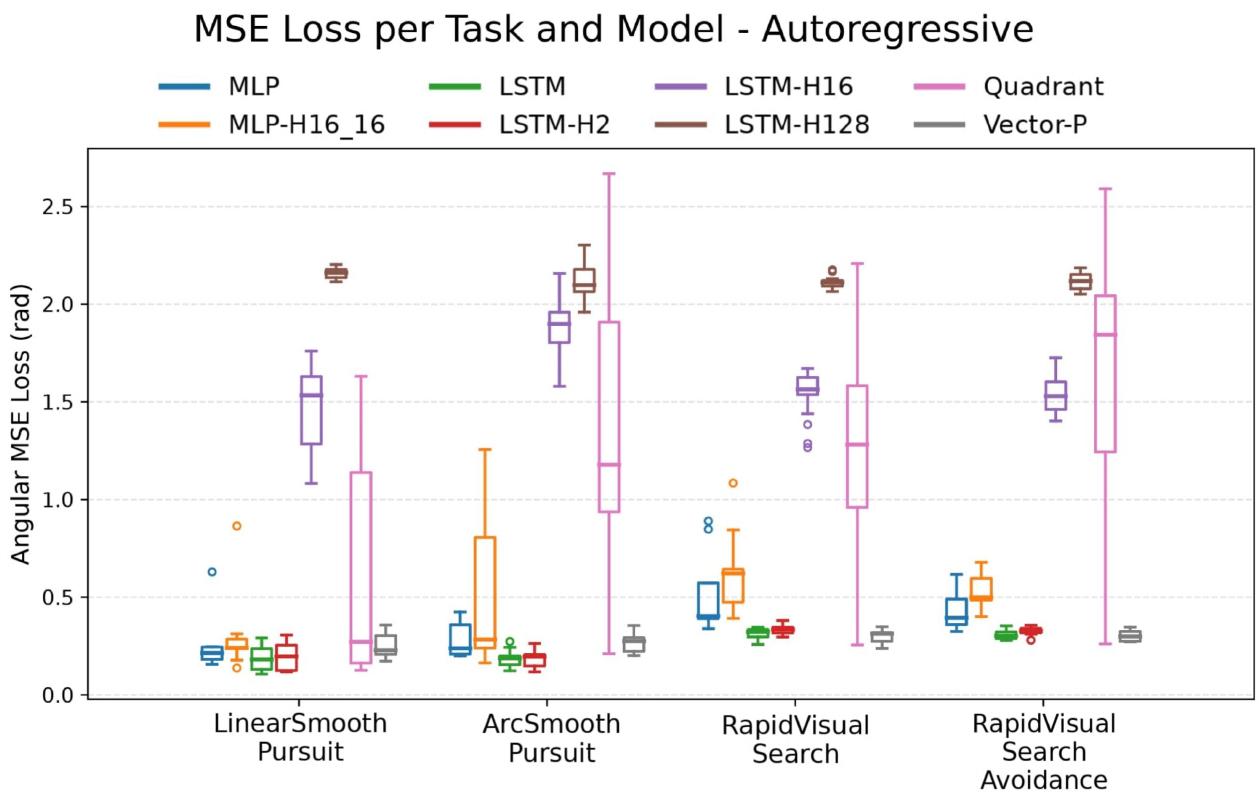}
    \caption{Distribution of mean squared error (lower is better) between imputed head direction and ground-truth head direction for various models over trajectories from the held-out test set. Results are grouped by task.}
    \label{fig:autoreg_result}
\end{figure}

Further, Fig.~\ref{fig:autoreg_result} compares mean squared error (MSE) between imputed and actual head directions, averaged over all trajectories, among various models. With these results, we can \textit{quantitatively} assess the viability of different configurations for each model. Models which have consistently low MSE scores across tasks can be deemed the most viable, whereas those with high variability or with consistently high scores are unlikely to be stable in VR, an inherently autoregressive environment.

Lastly, Table \ref{tab:autoreg_table} shows a comparison of model performance during training versus in the autoregressive experiment. The values in each row in the table represent the error of the corresponding model on the held-out test set and on the ground-truth head direction during the autoregressive procedure, respectively.

\begin{table}[htb]
    \caption{MSE Loss - Training and Autoregressive}
    \begin{center}
        \begin{tabularx}{0.8\linewidth}{lcc}
            \hline
            Model & Training & Autoregressive \\
            \hline
            MLP & 0.00011273 & 0.370922 \\
            MLP-H16\_16 & 0.00011108 & 0.499707 \\
            LSTM & 0.00032809 & 0.248416 \\
            LSTM-H2 & 0.00032678 & 0.261365 \\
            LSTM-H16 & 0.00037722 & 1.608792 \\
            LSTM-H128 & 0.00190842 & 2.125073 \\
            Quadrant (baseline) & 0.01127953 & 1.194305 \\
            Vector (parameterized) & 0.00011084 & 0.280166 \\
            \hline
        \end{tabularx}
    \end{center}
    \label{tab:autoreg_table}
\end{table}

\subsection{Discussion -- Autoregressive}
Our experiment shed light on which classes of controller may perform appropriately in VR and with the exoskeleton. Notably, as the size of the MLP and the hidden state within LSTM architectures increases, the stability of the head decreases under autoregressive conditions. For this reason, we selected an LSTM and MLP model with very small hidden states to be evaluated in the VR experiment. The Baseline Quadrant and Vector Parameterized controllers also moved forward to the VR experiment. Rejecting the four other controllers saves expensive time that is used instead for testing the more promising controllers.

\section{VR EXPERIMENT}
Having rejected four controllers through simulation, we carried the remaining four — Baseline Quadrant, Vector Parameterized, MLP, and LSTM — forward to the next evaluation layer. While the autoregressive results suggest these controllers behave stably in simulation, simulated rollouts cannot capture a real user's responses or preferences. We therefore deployed them in a VR environment, allowing human subjects to interact with each controller safely, before deploying them on the physical neck exoskeleton.

\subsection{Methods -- VR}

30 healthy subjects (21 male, 19-50 years old) volunteered to participate in the VR portion of the experiment. A healthy cohort allows us to prevent the unsafe controllers from reaching patients with DHS. 13 subjects had previous experience with VR; 19 wore prescription glasses daily; 5 participants wore glasses while wearing the VR headset. This experiment was approved by the Institutional Review Board (IRB) of The University of Utah. Informed consent was obtained from each participant prior to the experiment.

Our VR environment employs the open source code given by the eye-head coordination dataset paper~\cite{rubow2024dataset}, developed using the Unity game engine\footnote{\url{https://unity.com/}}. We used the VIVE Pro Eye VR headset with eye tracking technology\footnote{\url{https://www.vive.com/us/support/vive-pro-eye/category_howto/about-the-headset.html}}. This allowed us to mimic neck exoskeletons by moving the VR environment according to eye positions as if the subject was moving their head, even though their head was still.

\begin{figure}[htb]
    \centering
    \includegraphics*[width=0.7\linewidth]{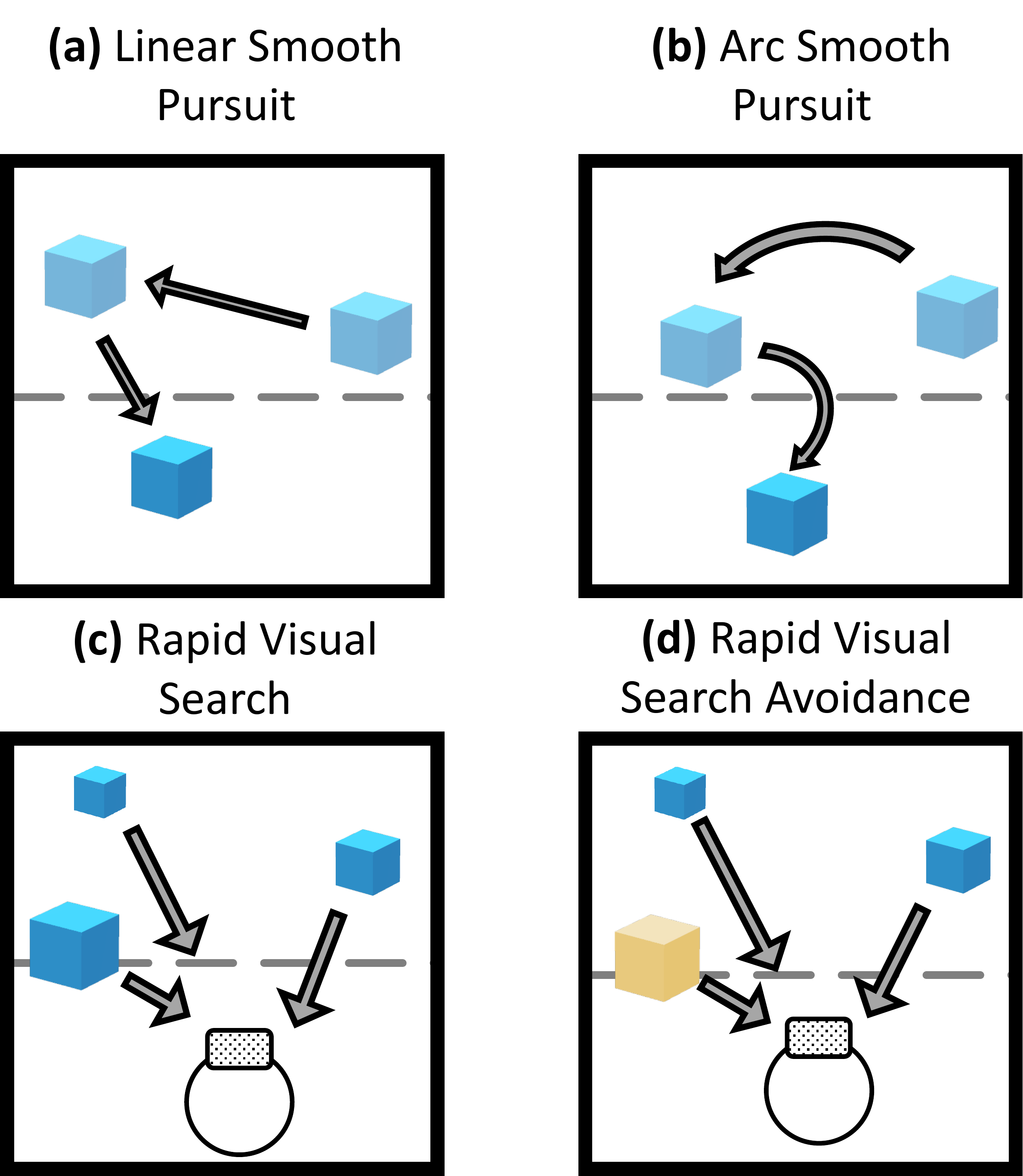}
    \caption{Example target trajectories. \textbf{(a)} The target in the \emph{Linear Smooth Pursuit} task moves in straight lines of random direction and distance. \textbf{(b)} The target in the \emph{Arc Smooth Pursuit} task moves in circular lines of random direction, distance, and curvature. Both pursuit tasks feature only a single target. \textbf{(c)} Each of the three targets in the \emph{Rapid Visual Search} task starts some distance from the participant and moves towards the participant. \textbf{(d)} In the \emph{Rapid Visual Search Avoidance} task there are three targets to fixate on (blue) and three blocking targets to not fixate on (yellow). Each target and blocking cube starts from a plane some distance from the participant and moves towards the participant. The dashed line represents a horizon. \label{fig:vr_tasks}}
\end{figure}

The four tasks (Fig.~\ref{fig:vr_tasks}), used to both create the dataset and perform this study, represent a diverse collection of eye-head behaviors. For all tasks, targets (cubic blocks) were constrained within a $140^\circ$ by $140^\circ$ conic range. When the gaze was fixed on a target, its color switched from blue to green.

The first two tasks were designed to measure and test head-eye tracking behaviors for a single, slow-moving target that either follows a sequence of random linear (\emph{Linear Smooth Pursuit}) or circular (\emph{Arc Smooth Pursuit}) paths. In the \emph{Linear Smooth Pursuit} task, a singular cube is instantiated in front of the user and moves in a straight line to a randomly assigned target point with a fixed speed (5 m/s). When reached, a new random target point is assigned from a uniform distribution, and the cube moves to the new point. The \emph{Arc Smooth Pursuit} task is similar, with the change that the cube moves in circular arc patterns with a fixed angular speed (1 rad/s).

The latter two tasks, \emph{Rapid Visual Search} and \emph{Rapid Visual Search Avoidance}, were intended to measure and test head-eye behavior during searching for discrete targets and brief fixation. In the \emph{Rapid Visual Search} task, three cubes are randomly instantiated within a $60^\circ$ head rotation of the user at a fixed distance (10~m). The cubes move towards the participant at a fixed speed. If the user fixates their gaze on a cube for 0.3 seconds or a cube reaches the user, the cube respawns at a new random location drawn from a uniform distribution. The \emph{Rapid Visual Search Avoidance} task is similar but also has distracting yellow cubes that turn orange on fixation. These distracting blocks do not disappear when fixated on and respawn when they reach the user.

Four controllers were selected after the simulation experiment, one from each category. The subject's goal was to perform the same VR tasks made for the eye-head coordination dataset~\cite{rubow2024dataset} with each controller and perform pairwise preference comparisons. For each task, the subjects first performed the task for 30 seconds without eye-tracking control to become familiar with the task. During this time, the subjects were able to explore the environment by physically moving their own head. Then, the subjects performed the task twice for two different controllers for 30 seconds each and chose their controller preference. During this time, the environment would move according to the subject's eye positions and not if the subject moved their head. The same process was repeated twice where preferred controllers moved forward to the next comparison for a total of three pairwise comparisons between four controllers. For controller comparison purposes, a point was given to the controller that was most preferred.

This comparison process was performed for each of the four tasks. Subjects were allowed to make no preference (i.e., both controllers were equally favored or disfavored) and, if no preference was chosen, one of the two controllers was randomly selected to move forward to the next comparison. If multiple controllers were equally preferred at the end of the task, the point was distributed evenly among all the preferred controllers. All pairwise orderings were randomly generated.

\subsection{Results -- VR}
\begin{figure}[htb]
    \centering
    \includegraphics*[width=\linewidth]{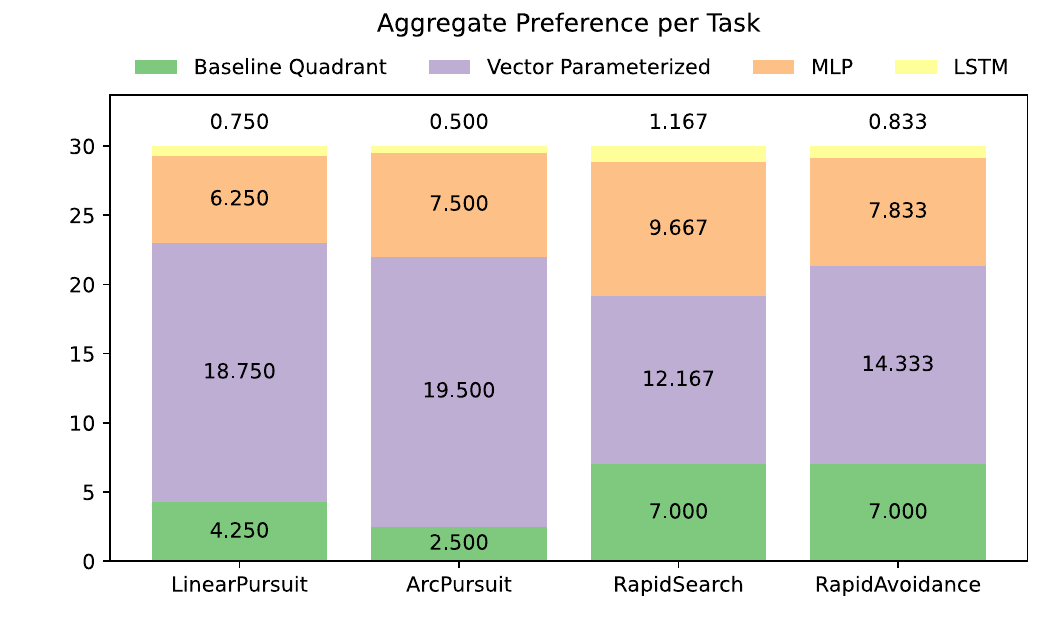}
    \caption{Stacked bar plot showing aggregate controller preference for each VR task. Since there were 30 subjects, the total number of points distributed across each controller was 30. A bar of greater length, or higher score, correlates to a more preferred controller for that task. \label{fig:vr_agg_per_task}}
\end{figure}

Since there were 30 subjects, for each of the four tasks, 30 points were distributed among the controllers. Fig.~\ref{fig:vr_agg_per_task} illustrates this. If for each task we rank the controllers according to their accumulated points, the ranking is the same for each task: Vector Parameterized is most preferred, followed by MLP, Baseline Quadrant, and LSTM, in that order. Because the LSTM model performed so poorly in terms of preference points as well as task score, it was rejected from the potential controller pool and barred from participating in the physical neck exoskeleton experiment.

\subsection{Discussion -- VR}
The MLP, Baseline Quadrant, and Vector Parameterized controllers have proven effective for performing tasks in a VR environment.

The LSTM model, while promising in an autoregressive simulation, proved nonviable and unwanted for restoring head-neck motions, challenging the hypothesis that previous eye and head positions are useful for predicting future head motions. We attribute the disparity in performance of the LSTM between training and VR to the problem known as teacher forcing~\cite{williams1989fullyrecurrent}. This occurs because, during training, predictions are discarded at each time step, whereas when used as an online controller, predictions affect the next state.

Employing VR as a tool for safely testing and rejecting bad or unsafe controllers has therefore saved more time for testing more promising controllers.

\section{NECK EXOSKELETON EXPERIMENT}
Even if a controller performs well in VR, this does not guarantee it will perform well in the physical robot. We therefore deployed and tested the three remaining controllers in a real-world environment with a physical neck exoskeleton.

\subsection{Methods -- Neck Exoskeleton}

\subsubsection{Neck Exoskeleton Technical Details}
\begin{figure}[htb]
    \centering
    \includegraphics*[width=0.8\linewidth]{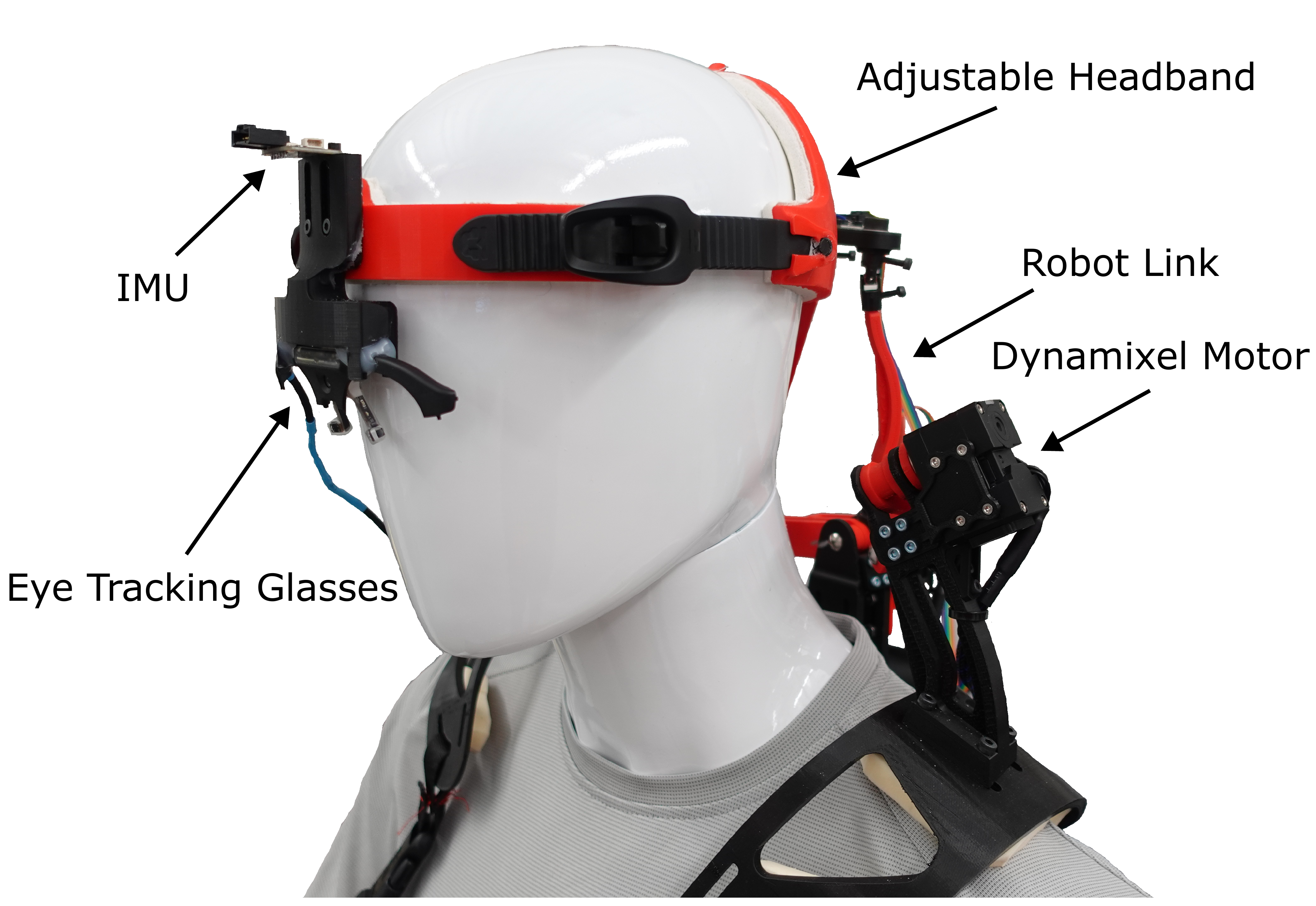}
    \caption{The Columbia Brace~\cite{zhang2017kinematic} neck exoskeleton. Links are manufactured from PLA 3D printed parts. Note that in this paper's implementation, the eye tracking glasses are not attached to the robot, but are worn directly by the user on the Neon glasses frame from Pupil Labs. \label{fig:neck_brace}}
\end{figure}

For the real-world robot experiment, we deployed our controllers on the Columbia Brace~\cite{zhang2017kinematic} (Fig.~\ref{fig:neck_brace}) whose design was optimized for maximizing range of motion and was thus a good fit for our healthy cohort. However, several improvements have been made to the exoskeleton.

For a smoother experience, velocity control was employed instead of position control with a maximum rotational speed of 1 radian per second. Software boundaries were set to prevent the robot from entering uncontrollable regions. Specifically, the robot was allowed $25^\circ$ flexion, $3^\circ$ extension, and $30^\circ$ side-to-side rotation in both directions. Roll or bending of the head was constrained to $0^\circ$. The robot was allowed to slide along its software boundaries.

To solve the forward kinematics problem, we used a neural network model as described in~\cite{prado2021artificial}. However, we increased the size of the model to 4 hidden layers of 200 nodes each and trained it on 28 million kinematic data points sampled uniformly to have a resolution of $0.1^\circ$ in each of the three head pose directions over the allowed robot workspace.

To measure eye angles, we used a commercial eye tracking system (Neon, Pupil Labs, Berlin, Germany) with an accuracy of $1.8^\circ$~\cite{baumann2023neon}. The system does not require calibration and is resistant to factors such as slipping, lighting, and blinking, and may be worn over glasses. The gaze positional data were sampled at 200~Hz, followed by filtering using an exponential moving average with a smoothing factor of 0.1.
Each of the controllers was run on a laptop CPU. Overall, the robotic system was operated at 50~Hz.

\subsubsection{Neck Exoskeleton Experiment}
All subjects from the VR evaluation experiment were invited, of whom twelve (7 male, 25-35 years old) agreed, to participate in the robotic evaluation follow-up. 
Signed informed consent was obtained prior to data collection, according to the IRB.

\begin{figure}[t]
    \centering
    \includegraphics*[width=\linewidth]{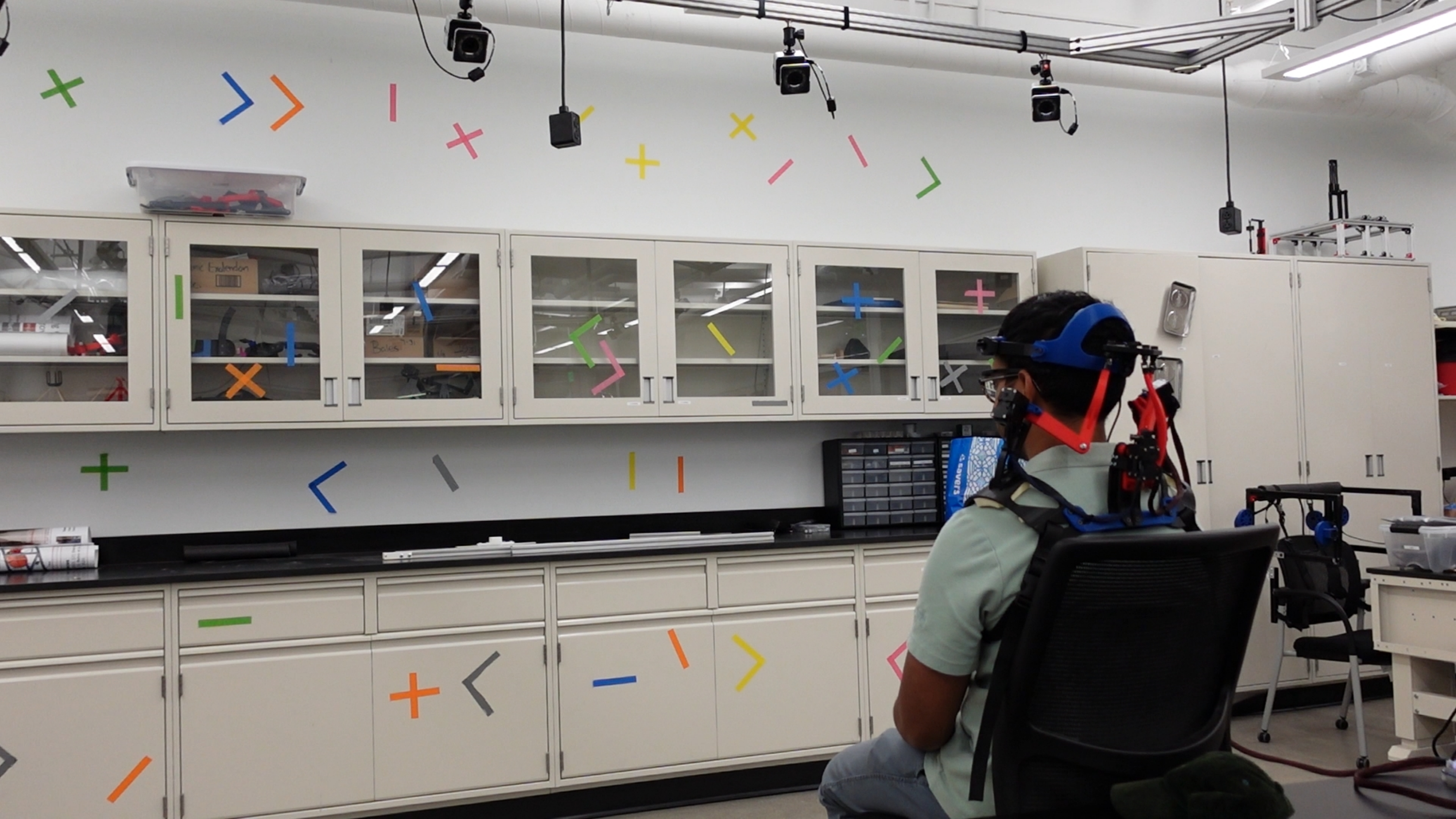}
    \caption{A subject provides preference regarding eye tracking controllers applied to the Columbia Brace~\cite{zhang2017kinematic}. The eye tracking frame is placed over the subject's glasses. The subject evaluated each controller by searching for a specific symbol and color combination on the wall. The clutter on the wall provides distraction to make the task more difficult and realistic. \label{fig:symbol_wall}}
\end{figure}

In this portion, subjects evaluated controllers during a real-world task. On a wall in front of the subject, 48 colored symbols were taped (Fig.~\ref{fig:symbol_wall}). The set of symbols was $\{X, +, \backslash, /, <, >, |, -\}$ and the set of colors was $\{$blue, yellow, pink, orange, green, silver$\}$. The task was to seek and find as many symbols as possible in 90 seconds. None of our subjects were colorblind, 10 wore glasses regularly, and 5 wore glasses during the experiment. Eye tracker accuracy was not appreciably affected by the presence of the glasses. 
This task modeled a real-world visual search for specific targets among many distracting objects. The targets spanned $90^\circ$ evenly across the subjects' field of view. 

Users were given 5 minutes to become comfortable controlling the neck exoskeleton. Afterward, the subject followed an experimental protocol similar to the VR experiment. Each subject performed the task with two of the controllers (MLP, Vector, or baseline) in turn and then stated a preference for either of them. The subject then performed the task with the preferred controller and the remaining controller and stated a preference for either of those. A `no preference' option was allowed, and all orderings were randomized. Subjects were blinded to which controller they were using. If a subject expressed no preference between controllers in the first comparison, each controller was considered equally preferred, and the controller that moved forward was randomly selected.

\subsection{Results -- Neck Exoskeleton}
The task scores and preferences for each controller for the neck exoskeleton experiment are reported in Fig.~\ref{fig:score_pref_robot}. By normalizing each subject's score and averaging the score of the repeated controller, we found that the mean normalized score (standard deviation) for the Baseline Quadrant, Vector Parameterized, and MLP controllers respectively are 0.575(0.068), 0.543(0.092), and 0.596(0.078). Each controller thus performed similarly well on average. Three subjects preferred the Vector Parameterized controller, four preferred the MLP, and five preferred the Baseline Quadrant. Overall, the Baseline Quadrant controller was the most preferred in this experiment, but all controllers passed the evaluation. Five of the subjects gave preference to their highest scoring controller, while the rest preferred controllers that underperformed compared to the other controllers.

\newcommand{\cquadrant}[1]{\cellcolor{green!#1!white}}
\newcommand{\cvector}[1]{\cellcolor{blue!#1!white}}
\newcommand{\cmlp}[1]{\cellcolor{orange!#1!white}}
\newcommand{\win}[1]{\textbf{#1}}
\newcolumntype{Y}{>{\centering\arraybackslash}X}
\begin{figure}[t]
    \begin{center}
        \begin{tabularx}{\linewidth}{rcYYcYY}
        \cquadrant{40}Quadrant & & \multicolumn{2}{c}{\cvector{40}Vector} & & \multicolumn{2}{c}{\cmlp{40}MLP} \\ \\
        subject & & \multicolumn{2}{c}{1st comparison} & & \multicolumn{2}{c}{2nd comparison} \\
        \cmidrule{1-1}\cmidrule{3-4}\cmidrule{6-7}
        0 & &\cvector{21}\win{11} & \cquadrant{32}17 & & \cmlp{26}14 & \cvector{21}\win{11} \\
        1 & & \cmlp{20}13 & \cquadrant{28}\win{18} & & \cvector{29}19 & \cquadrant{23}\win{15} \\
        2 & & \cquadrant{23}10 & \cvector{9}\win{4} & & \cvector{30}13 & \cmlp{37}\win{16} \\
        3 & & \cmlp{20}\win{6} & \cvector{27}8 & & \cmlp{27}8 & \cquadrant{27}\win{8} \\
        4 & & \cvector{22}\win{16} & \cquadrant{26}19 & & \cvector{25}18 & \cmlp{27}\win{20} \\
        5 & & \cmlp{23}13 & \cquadrant{27}\win{15} & & \cquadrant{27}15 & \cvector{23}\win{13} \\
        6 & & \cquadrant{24}\win{14} & \cvector{28}\win{16} & & \cquadrant{22}13 & \cmlp{26}\win{15} \\
        7 & & \cmlp{30}\win{18} & \cvector{20}12 & & \cquadrant{27}\win{16} & \cmlp{23}14 \\
        8 & & \cquadrant{19}\win{9} & \cvector{30}14 & & \cmlp{23}11 & \cquadrant{28}\win{13} \\
        9 & & \cmlp{28}\win{9} & \cvector{19}6 & & \cmlp{25}8 & \cquadrant{28}\win{9} \\
        10 & & \cquadrant{19}9 & \cvector{28}\win{13} & & \cmlp{28}13 & \cvector{26}\win{12} \\
        11 & & \cquadrant{21}\win{10} & \cvector{21}10 & & \cquadrant{28}13 & \cmlp{30}\win{14} \\
        \end{tabularx}
    \end{center}
    \caption{Task score and preference rankings for neck exoskeleton experiment. The colored cells represent a trial using a particular controller; green for Baseline Quadrant, blue for Vector Parameterized model, and orange for MLP. Reading from left to right corresponds to the order each controller was given to the subjects. The score (number of found symbols) is printed on each cell which is colored with a shading that corresponds to the normalized score for that subject. Thus, for a row, a darker shaded cell corresponds to a higher score. A bold score indicates the controller was preferred in that comparison. If both scores are bold then equal preference was reported. Overall, three subjects preferred Vector Parameterized, four preferred MLP, and five preferred Baseline Quadrant. \label{fig:score_pref_robot}}
\end{figure}

\begin{figure}[htb]
    \centering
    \includegraphics*[width=\linewidth]{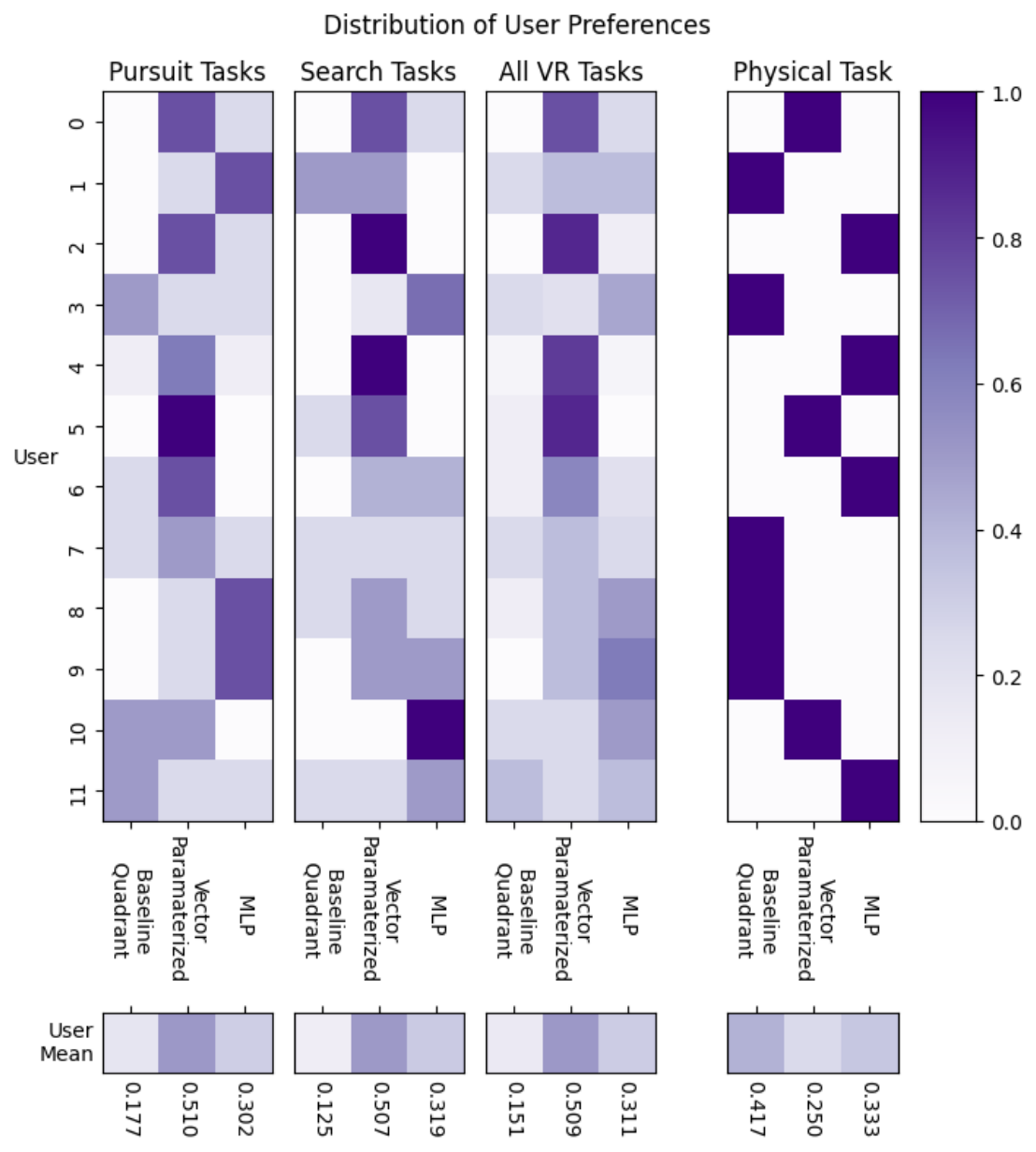}
    \caption{User preferences per user and averaged over all users. Darker shading indicates a more preferred controller. \emph{Linear} and \emph{Arc Pursuit} were grouped together into Pursuit Tasks because they are similar. \emph{Rapid Search} and \emph{Rapid Avoidance} were also grouped together due to their similar natures. The preference over all tasks is also present in the third column and the overall controller preference during the physical neck exoskeleton experiment is present in the fourth column. Preference averaged over all subjects is presented in the bottom row. \label{fig:pref_dist}}
\end{figure}

When comparing overall controller preferences in the VR experiment and the Neck Exoskeleton experiment, shown in Fig.~\ref{fig:pref_dist}, preferences were in agreement for three subjects. If we perform the same comparison for the first two VR tasks, the tracking pursuit tasks, preferences were in agreement for four subjects. The same comparison for the final two VR tasks, the search and fixate tasks, shows that preferences were in agreement for six subjects. This matches our intuition that, because the search and fixate tasks are the most similar to the real-world neck exoskeleton task, we should see that preferences over controllers are most aligned between the VR environments and the real-world exoskeleton environments in these settings. In the separate bottom row of Fig.~\ref{fig:pref_dist}, we again see that the Vector Parameterized model was the most preferred controller in the VR experiment and the Baseline Quadrant was the most preferred controller in the Neck Exoskeleton experiment. 

\subsection{Discussion -- Neck Exoskeleton}
Three controllers passed the VR evaluation, and all three also passed the neck exoskeleton evaluation. Employing our models in VR first has thus allowed us to find viable controllers before deployment on a physical robot.

We found that for VR tasks similar to the real-world task, half of the subjects preferred the same controller. Furthermore, we found that while the Baseline Quadrant controller was preferred the least of the three final controllers in VR, it was preferred the most with the physical neck exoskeleton. When making preference, subjects weighed task performance less than other values such as comfort or safety.

The sim-to-real gap between VR and the real world partially accounts for differences of controller preference. The physical neck exoskeleton exerts physical forces on the user's head. The controllers in VR do not account for this dynamic. The sim-to-real gap may also account for the extra preference for the Baseline Quadrant controller on the neck exoskeleton. In the VR environment, the Baseline Quadrant controller could only move in strict horizontal and vertical directions, which may feel jarring. The controller behaves the same algorithmically on the neck exoskeleton, but may feel less jarring due to the robot's dynamics, including damping. Indeed, one of our subjects verbally expressed that the Baseline Quadrant controller felt smooth.

Subjects also expressed that the Vector Parameterized controller moved too fast, was harder to control horizontally, and still moved slightly when fixating on a target. It was expressed that the MLP controller was harder to control vertically. The Baseline Quadrant controller may be preferred more often because it is simple and predictable.

The results of our physical experiment provide evidence that we have found two more viable controllers beyond the Baseline Quadrant that may serve the purpose of restoring head motions. Because there is a large amount of disagreement among user preferences, our results suggest that personalization is an important factor in gaze-driven controllers, motivating the need for multiple viable controllers.

\section{CONCLUSION}
We demonstrated a novel multi-layer design and evaluation framework that leverages VR to safely and efficiently test for viable neck exoskeleton controllers before physical deployment, saving significant development time and resources. As a result, we developed and verified two novel gaze-driven eye-tracking controllers for restoring head-neck motion. 
Interestingly, our user study results reveal that no single controller is universally preferred: while some subjects favored a simple baseline, others preferred our novel data-driven models. These findings highlight the importance of personalization in assistive robotics~\cite{he2023learning} and motivate the need for future work on learning the preferences~\cite{liu2023efficient} and implicit  constraints~\cite{papadimitriou2024bayesian} of neck exoskeleton users. 



\bibliographystyle{IEEEtran}
\bibliography{sources}

\end{document}